%% file: 00-main.tex
\documentclass[final]{cvpr}

\usepackage{times}
\usepackage{epsfig}
\usepackage{graphicx}
\usepackage{amsmath}
\usepackage{amssymb}
\usepackage{amsfonts}
\usepackage{amsthm}
\usepackage{url}
\usepackage{color}
\usepackage{bbm}
\usepackage{tabularx}
\usepackage{xspace}
\usepackage{paralist} %
\usepackage{caption}
\usepackage{booktabs}
\usepackage{microtype}
\usepackage{wrapfig}
\usepackage{adjustbox}
\usepackage{multirow}
\usepackage{afterpage}
\usepackage{makecell}
\usepackage[font=small]{caption}
\usepackage[outline]{contour}

\let\originalleft\left
\let\originalright\right
\renewcommand{\left}{\mathopen{}\mathclose\bgroup\originalleft}
\renewcommand{\right}{\aftergroup\egroup\originalright}

\newcommand{\norm}[1]{\left\lVert#1\right\rVert}

\usepackage[pagebackref=true,breaklinks=true,colorlinks,bookmarks=false]{hyperref}

\def\cvprPaperID{4114} %
\def\confYear{CVPR 2021}

\begin{document}

\title{IBRNet: Learning Multi-View Image-Based Rendering}

\author{
Qianqian Wang$^{1, 2}$  \ \ \ 
Zhicheng Wang$^1$ \ \ \
Kyle Genova$^{1, 3}$ \ \ \
Pratul Srinivasan$^1$ \ \ \
Howard Zhou$^1$ \and
Jonathan T. Barron$^1$ \ \ \ 
Ricardo Martin-Brualla$^1$ \ \ \
Noah Snavely$^{1,2}$ \ \ \
Thomas Funkhouser$^{1,3}$
\\[2mm]
\vspace{1em}
$^1$Google Research \ \ \
$^2$Cornell Tech, Cornell University \ \ \
$^3$Princeton University
}

\maketitle

\begin{abstract}
We present a method that synthesizes novel views of complex scenes by interpolating a sparse set of nearby views. The core of our method is a network architecture that includes a multilayer perceptron and a ray transformer that estimates radiance and volume density at continuous 5D locations~(3D spatial locations and 2D viewing directions), drawing appearance information on the fly from multiple source views. By drawing on source views at render time, our method hearkens back to classic work on image-based rendering (IBR), and allows us to render high-resolution imagery. Unlike neural scene representation work that optimizes per-scene functions for rendering, we learn a generic view interpolation function that generalizes to novel scenes. We render images using classic volume rendering, which is fully differentiable and allows us to train using only multi-view posed images as supervision. Experiments show that our method outperforms recent novel view synthesis methods that also seek to generalize to novel scenes. Further, if fine-tuned on each scene, our method is competitive with state-of-the-art single-scene neural rendering methods.\footnote{\url{https://ibrnet.github.io/}}

\end{abstract}

\input{01-intro}

\input{02-related}
\input{03-method}

\input{04-experiment}

\input{05-conclusion}

{\small
\bibliographystyle{ieee_fullname}
\bibliography{refs}
}

\input{supp}

\end{document}

%% file: 01-intro.tex
\section{Introduction}
Given a set of posed images of a scene, the goal of novel view synthesis is to produce photo-realistic images of the same scene at novel viewpoints. 
Early work on novel view synthesis focused on image-based rendering~(IBR).
Starting from the pioneering work on view interpolation of Chen and Williams~\cite{chen1993view}, and proceeding through light field rendering~\cite{buehler2001unstructured,gortler1996lumigraph, levoy1996light}, view-dependent texturing~\cite{debevec1998efficient}, and more modern learning-based methods~\cite{hedman2018deepblending}, IBR methods generally operate by warping, resampling, and/or blending source views to target viewpoints. Such methods can allow for high-resolution rendering, but generally require either very dense input views or explicit proxy geometry, which is difficult to estimate with high quality leading to artifacts in rendering.

More recently, one of the most promising research directions for novel view synthesis is neural scene representations, which represent scenes as the weights of neural networks. This research area has seen significant progress through the use of Neural Radiance Fields~(NeRF)~\cite{nerf}.
NeRF shows that multi-layer perceptrons~(MLPs) combined with positional encoding can be used to represent the continuous 5D radiance field of a scene, enabling photo-realistic novel view synthesis on complex real-world scenes.
NeRF's use of continuous scene modeling via MLPs, as opposed to explicit discretized volumes~\cite{sitzmann2019deepvoxels} or multi-plane images~\cite{flynn2019deepview,zhou2018stereo} allows for more compact representations and scales to larger viewing volumes.

Although neural scene representations like NeRF can represent scenes faithfully and compactly, they typically require a lengthy optimization process for each new scene before they can synthesize any novel views of that scene, which limits the value of these methods for many real-world applications.

In this work, we leverage ideas from both IBR and NeRF into a new learning-based method that generates a \emph{continuous} scene radiance field on-the-fly from multiple source views for rendering novel views.
We learn a general view interpolation function that simultaneously performs density/occlusion/visibility reasoning and color blending while rendering a ray. 
This enables our system to operate without any scene-specific optimization or precomputed proxy geometry.

At the core of our method is a lightweight MLP network that we call \emph{IBRNet}, which aggregates information from source views along a given ray to compute its final color. 
For sampled 3D locations along the ray, the network first fetches latent 2D features, derived from nearby source views, that encode spatial context. 
IBRNet then aggregates these 2D features for each sampled location to produce a \emph{density feature} that captures information about whether that feature seems to be on a surface.
A ray transformer module then computes a scalar density value for each sample by considering these density features 
along the entire ray, enabling visibility reasoning across larger spatial scales.
Separately, a color blending module uses the 2D features and view direction vectors from source views to derive a view-dependent color for each sample, computed as a weighted combination of the projected colors of the source views.
A final color value is then computed for each ray using volume rendering. 

Our approach is fully differentiable and can therefore be trained end-to-end using multi-view images. 
Our experiments show that when trained on large amounts of data, our method can render high-resolution photo-realistic novel views for unseen scenes that contain complex geometry and materials, and our quantitative evaluation shows that it improves upon state-of-the-art novel view synthesis methods designed to generalize in a single shot to new test scenes.
Moreover, for a particular scene, we can fine-tune IBRNet to improve the quality of synthesized novel views to match the performance of state-of-the-art neural scene representation methods like NeRF~\cite{nerf}.
In summary, our contributions are:
\begin{compactenum}[--]
\item 
a new learning-based multi-view image-based rendering approach that outperforms existing one-shot view synthesis methods on novel scenes,
\item 
a new model architecture called IBRNet that enables the continuous prediction of colors and densities in space from multiple views,
\item 
a per-scene fine-tuning procedure that achieves comparable performance to state-of-the-art novel view synthesis methods designed only for single-scene inference.
\end{compactenum}

%% file: 02-related.tex
\section{Related work}

\noindent\textbf{Image based rendering.}
Early work on IBR introduced the idea of synthesizing novel views from a set of reference images by a weighted blending of reference pixels \cite{debevec1996modeling,gortler1996lumigraph,levoy1996light}.  Blending weights were computed based on ray-space proximity \cite{levoy1996light} or approximate proxy geometry \cite{buehler2001unstructured,debevec1996modeling,heigl1999plenoptic}.   In more recent work, researchers have proposed improved methods for computing proxy geometry \cite{chaurasia2013depth,hedman2016scalable}, optical flow correction \cite{casas20154d,du2018montage4d,eisemann2008floating}, and soft blending \cite{penner17,Riegler2020FVS}. For example, Hedman et al.~\cite{hedman2018deepblending} use two types of multi-view stereo \cite{jancosek2011multi,schonberger2016structure} to produce a view-dependent mesh surface, then use a CNN to compute blending weights. Others synthesize a radiance field directly on a mesh surface  \cite{debevec1998efficient,huang2020adversarial,thies2019deferred} or point cloud~\cite{aliev2019neural, meshry2019neural,pittaluga2019revealing}.  While these methods can handle sparser views than other approaches and achieve promising results in some cases, they are fundamentally limited by the performance of 3D reconstruction algorithms~\cite{jancosek2011multi,schonberger2016structure}.  They have difficulty in low-textured or reflective regions, where stereo reconstruction tends to fail, and cannot handle partially translucent surfaces. In contrast, our method learns continuous volume densities in an end-to-end manner that is optimized for synthesis quality, 
leading to better performance in challenging scenarios.

\medskip
\noindent \textbf{Volumetric Representations.} Another line of work uses discrete volumetric representations to achieve photo-realistic rendering. 
Recent methods leverage convolutional neural networks~(CNNs) to predict volumetric representations stored in voxel grids \cite{henzler20,kalantari2016learning,kar17,penner17,tulsiani17} or multi-plane images~(MPIs)~\cite{flynn2019deepview,flynn2016deepstereo,li2020crowdsampling,mildenhall19,srinivasan2019pushing,zhou2018stereo}. At test time, novel views can be rendered from these representations via alpha compositing~\cite{porter1984compositing}. Trained end-to-end on large datasets, these methods often learn to compensate for the discretization artifacts of low-resolution voxel grids and can generalize reasonably well to different scenes. While they achieve high-quality view synthesis results, they must explicitly process and store large 
numbers of samples resulting in extensive memory overhead, limiting the resolution of their outputs. 
In contrast, our method allows for querying color and opacity at continuous 3D locations and 2D viewing directions without storing a full scene representation, and 
can scale to render high-resolution images. 
Our method can also handle larger viewing volumes than MPI-based methods.

\medskip
\noindent \textbf{Neural scene representations.}
A promising recent research direction 
is the use of neural networks for representing the shape and appearance of scenes.
Earlier work \cite{atzmon2019controlling,genova2020,jiang2020, occnet,deepsdf,Peng2020ECCV,disn} showed that MLPs can be used as implicit shape representations, where the weights of the MLPs map continuous spatial coordinates to signed distance or occupancy values. With advances in differentiable rendering methods~\cite{chen2019dibrender,jiang2020sdfdiff,redner,liu2019soft,liu2020dist,DVR}, many methods~\cite{nsvf,neuralvolumes,DVR,saito2019pifu,schwarz2020graf,sitzmann2019deepvoxels,srn,tancik2020fourier,yariv2020multiview} showed the ability to learn both scene geometry and appearance from multi-view observations for simple geometry and diffuse materials. The more recent NeRF~\cite{nerf} achieves very impressive results for novel view synthesis by optimizing a 5D neural radiance field for a scene, 
suggesting the advantages of continuous representations over discrete ones like voxel grids or meshes. While NeRF opens up many new research opportunities~\cite{li2020neural,martinbrualla2020nerfw,park2020nerfies,peng2021neural,Schwarz2020NEURIPS, nerv2020}, it must be optimized for each new scene, taking 
hours or days to converge. Concurrent work~\cite{trevithick2020grf,yu2020pixelnerf} tries to address this issue and generalize NeRF~(with a focus on very sparse input views). However, 
their use of absolute locations as direct network inputs restricts their ability to generalize to arbitrary new scenes.
In contrast, our method can generalize to new, real-world scenes with high quality.

%% file: 03-method.tex
\section{Method}
\begin{figure*}[t]
    \centering
    \includegraphics[width=\linewidth]{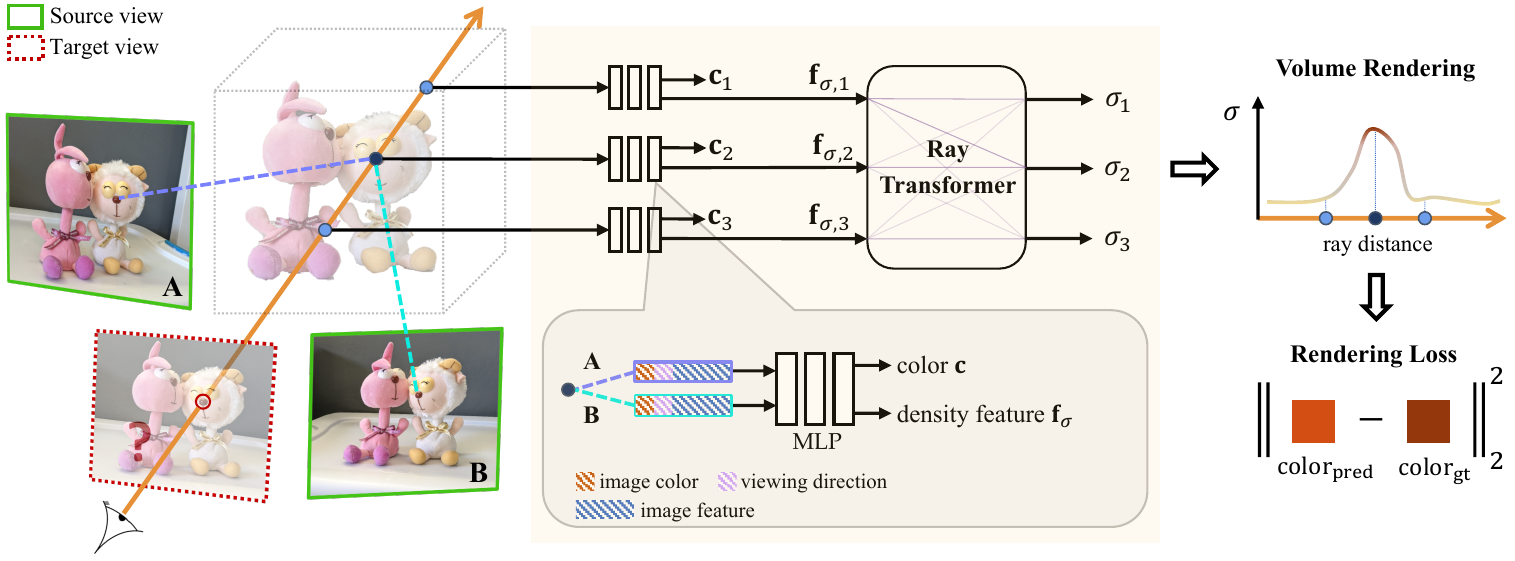}
    \caption{
    \textbf{System Overview}.
    1) To render a novel target view (shown here as the image labeled with a `?'), we first identify a set of neighboring source views (e.g., the views labeled \textbf{A} and \textbf{B}) and extract their image features. 2) Then, for each ray in the target view, we compute colors and densities for a set of samples along the ray using our proposed IBRNet~(yellow shaded region). Specifically, for each sample, we aggregate its corresponding information~(image colors, features, and viewing directions) from the neighboring source views to produce its color $\mathbf{c}$ and density features $\mathbf{f}_{\sigma}$~(note that these features are not yet scalar density values). We then apply our proposed ray transformer to these density features across all samples on the ray to predict densities.  
    3) Finally, we use volume rendering to accumulate colors and densities along the ray to render its color. Our method can be trained end-to-end with an $L_2$ loss on reconstructed image colors.
    }
    \label{fig:overview}
\end{figure*}
Given nearby source views, our method uses volume rendering to synthesize a target view at a novel camera pose. The core problem we try to solve is to obtain colors and densities in a continuous space by aggregating information present in the source views. Our system pipeline (Fig.~\ref{fig:overview}) can be divided into three parts:
1) identifying a set of nearby source views as input and extracting dense features from each source view,
2) predicting volume densities~$\sigma$ and colors~$\mathbf{c}$ at continuous 5D locations~(3D spatial locations and 2D view directions), and 
3) compositing those colors and densities along each camera ray through volume rendering to produce a synthesized image.

\subsection{View selection and feature extraction}
Unlike neural scene representations that attempt to encode an entire scene into a single network, we synthesize the novel target view by interpolating nearby source views.
While our network (described in the next section) can handle an arbitrary number of neighboring views, given limited GPU memory we select a small number of source views as the ``working set'' for rendering a novel view.
To obtain an effective working set, we identify spatially nearby candidate views, then select the subset of $N$ views whose viewing directions are most similar to the target view. 

Let $\mathbf{I}_i\in[0,1]^{H_i\times W_i\times 3}$ and $\mathbf{P}_i\in\mathbb{R}^{3\times4}$ respectively denote the color image and camera projection matrix for the $i$-th source view. We use a shared U-Net based convolutional neural network to extract dense features $\mathbf{F}_i\in\mathbb{R}^{H_i\times W_i\times d}$ from each image~$\mathbf{I}_i$. 
The set of tuples $\{(\mathbf{I}_i, \mathbf{P}_i, \mathbf{F}_i)\}_{i=1}^N$ forms the input for rendering a target view.

\begin{figure*}[t]
    \centering
    \includegraphics[width=0.8\linewidth]{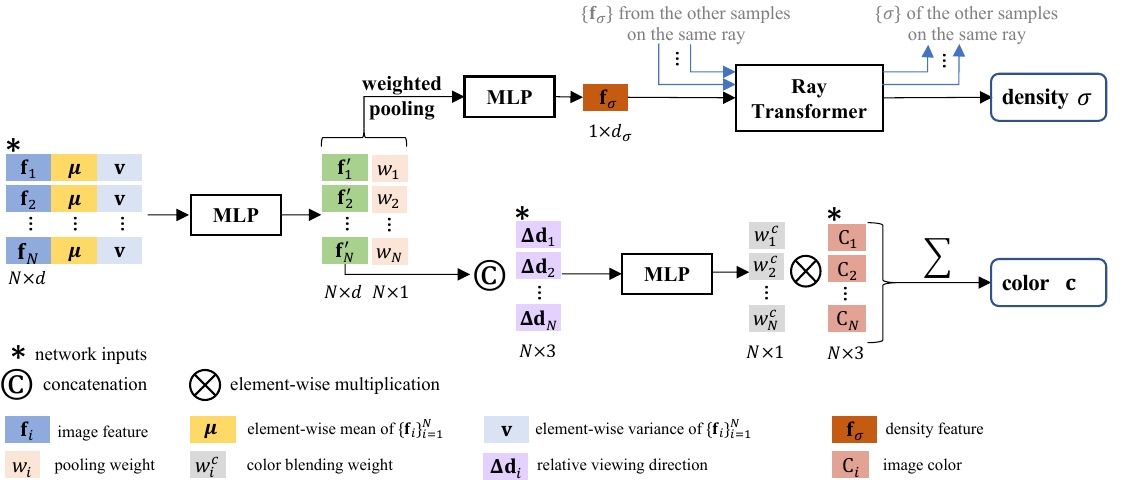}
    \caption{
    \textbf{IBRNet for volume density and color prediction at a continuous 5D location~$(\mathbf{x}, \mathbf{d})$.}
    We first input the 2D image features $\{\mathbf{f}_i\}_{i=1}^N$ extracted from all source views to a PointNet-like MLP to aggregate local and global information, resulting in multi-view aware features $\{\mathbf{f}'_i\}_{i=1}^N$ and pooling weights $\{w_i\}_{i=1}^N$. To predict density, we pool $\{\mathbf{f}'_i\}_{i=1}^N$ using weights $\{w_i\}_{i=1}^N$ which enables multi-view visibility reasoning to obtain a density feature $\mathbf{f}_\sigma$. Instead of directly predicting density $\sigma$ from $\mathbf{f}_\sigma$ for individual 5D samples, we use a \emph{ray transformer} module to aggregate information of all samples along the ray. The ray transformer module takes $\mathbf{f_\sigma}$ for all samples on a ray and predicts all their densities~(only the density output for $(\mathbf{x}, \mathbf{d})$ is highlighted in the figure for simplicity). The ray transformer module enables geometric reasoning across a longer range and improves density predictions. For color prediction, we concatenate $\{\mathbf{f}'_i\}_{i=1}^N$ with the viewing directions of the query ray relative to each viewing direction of the source view, i.e., $\{\Delta\mathbf{d}_i\}_{i=1}^N$, and predict a set of blending weights. The output color~$\mathbf{c}$ is a weighted average of the image colors from the source views.
    }
    \label{fig:method}
\end{figure*}

\subsection{RGB-$\sigma$ prediction using IBRNet}
Our method synthesizes images based on classic volume rendering, accumulating colors and densities in the 3D scene to render 2D images. 
We propose IBRNet~(illustrated in Fig.~\ref{fig:method}) to predict colors and densities at continuous 5D locations by aggregating information from multiple source views and incorporating long-range context along the ray. Our proposed IBRNet is permutation-invariant and accepts a variable number of source views. 

We first describe the input to IBRNet for a single query point location $\mathbf{x}\in\mathbb{R}^3$
on a ray~$\mathbf{r} \in\mathbb{R}^3$ with unit-length viewing direction~$\mathbf{d} \in\mathbb{R}^3$.
We project $\mathbf{x}$ into all source views using their camera parameters, and extract colors and features at the projected pixel locations through bilinear interpolation. Let $\{C_i\}_{i=1}^N\in [0,1]^3$ and $\{\mathbf{f}_i\}_{i=1}^N \in \mathbb{R}^d$ respectively denote these extracted colors and image features for point $\mathbf{x}$.
We also take into account the viewing directions for $\mathbf{x}$ in all source views, denoted as $\{\mathbf{d}_i\}_{i=1}^N$. 

\subsubsection{Volume density prediction}
Our density prediction at point $(\mathbf{x}, \mathbf{d})$ involves two steps. First, we aggregate the multi-view features at $(\mathbf{x}, \mathbf{d})$ to obtain a density feature. Then, our proposed ray transformer takes in density features for all samples on a ray and incorporates long-range contextual information to predict the density for each sample including $(\mathbf{x}, \mathbf{d})$.

\medskip 
\noindent \textbf{Multi-view feature aggregation.}
We observe that 3D points on surfaces are more likely to have consistent local appearances in multiple views than 3D points in free space. 
Therefore, an effective way to infer density would be to check the consistency among the features $\{\mathbf{f}_i\}_{i=1}^N$ for a given point, and a possible implementation would be a PointNet-like~\cite{qi2016pointnet} architecture that takes in multi-view features and uses variance as the global pooling operator. 
Specifically, we first compute a per-element mean~$\boldsymbol{\mu}\in\mathbb{R}^d$ and variance~$\mathbf{v}\in\mathbb{R}^d$ from features $\{\mathbf{f}_i\}_{i=1}^N$ to capture global information, and concatenate each $\mathbf{f}_i$ with $\boldsymbol{\mu}$ and $\mathbf{v}$. Each concatenated feature is fed into a small shared MLP to integrate both local and global information, resulting in a multi-view aware feature $\mathbf{f}_i'$ and a weight vector~$w_i\in[0, 1]$. We pool these new features $\{\mathbf{f}_i'\}_{i=1}^N$ by computing their weighted average and variance using weights $\{w_i\}_{i=1}^N$, which are mapped to a density feature $\mathbf{f}_\sigma\in\mathbb{R}^{d_\sigma}$ using an MLP. Compared to a direct average or max-pooling in a PointNet~\cite{qi2016pointnet}, we find that our weighted pooling improves the network's ability to handle occlusions.

\medskip
\noindent \textbf{Ray transformer.}
After obtaining a density feature $\mathbf{f}_\sigma$, one could directly turn it into a single density $\sigma$ with another MLP. However, we find such an approach fails to predict accurate densities for new scenes with complex geometry. 
We ascribe this to the fact that looking at features for a single point \emph{in isolation} is inadequate for accurately predicting its density, and more contextual information is needed---similar to how plane-sweep stereo methods consider matching scores along a whole ray before determining the depth of a particular pixel.

We therefore introduce a new ray transformer module to enable samples on a ray to attend to each other before predicting their densities. The ray transformer consists of the two core components of a classic Transformer~\cite{vaswani2017attention}: positional encoding and self-attention. Given $M$ samples along the ray, our ray transformer treats samples from near to far as a sequence, and applies positional encoding and multi-head self-attention to the sequence of density features $(\mathbf{f}_\sigma(\mathbf{x}_1), \cdots, \mathbf{f}_\sigma(\mathbf{x}_M))$.  The final density value~$\sigma$ for each sample is then predicted from its attended feature.
The ray transformer module allows for a variable number of inputs and only introduces only a small increase in 
parameters and computational overhead, while
dramatically improving the quality of the predicted densities and the final synthesized image, as shown in our ablation studies.

\medskip
\noindent \textbf{Improving temporal visual consistency.} 
Our method considers only nearby source views as the working set when synthesizing a target view. Therefore, when generating videos along smooth camera paths, our method is potentially subject to temporarily inconsistent density predictions and flickering artifacts due to abrupt changes in the working set as the camera moves.

To alleviate this issue, we adopt the pooling technique of Sun \etal~\cite{sun2020light}. We replace the mean $\boldsymbol{\mu}$ and variance $\mathbf{v}$ of $\{\mathbf{f}_i\}_{i=1}^N$ with a weighted mean $\boldsymbol{\mu}_w$ and variance $\mathbf{v}_w$, weighted so as to reduce the influence of the furthest images in the working set.
The weight function is defined as:
\begin{align}
    & \Tilde{w}^{\mathbf{f}}_i(\mathbf{d}, \mathbf{d}_i) = \max\left(0, e^{s(\mathbf{d}\cdot \mathbf{d}_i -1)} - \underset{j=1...N}{\min}e^{s(\mathbf{d}\cdot \mathbf{d}_j -1)}\right)\,, \nonumber \\
    & w^{\mathbf{f}}_i(\mathbf{d}, \mathbf{d}_i) = \frac{\Tilde{w}^{\mathbf{f}}_i(\mathbf{d}, \mathbf{d}_i)}{\sum_j \Tilde{w}^{\mathbf{f}}_j(\mathbf{d}, \mathbf{d}_j)}\,,
\label{eq:pooling_weight}
\end{align}
where $s$ is a learnable parameter.
This technique improves our method in two ways: 1)
it smooths the change in global features between adjacent frames, and 2) it produces more reasonable global features by up-weighting closer views in the working set and down-weighting more distant views. We observe this technique empirically improves synthesis stability and quality.

\subsubsection{Color prediction}
We obtain color at a 5D point by predicting blending weights for the image colors~$\{C_i\}_{i=1}^N$ in the source views that correspond to that 5D point. 
Unlike NeRF~\cite{nerf}, which uses absolute viewing directions, we consider viewing direction relative to that of the source views, i.e., the difference between $\mathbf{d}$ and $\mathbf{d}_i$. A smaller difference between $\mathbf{d}$ and $\mathbf{d}_i$ typically implies a larger chance of the color at target view resembling the corresponding color at view $i$, and vice versa.

To predict the blending weight for each source color~$C_i$, we concatenate the feature $\mathbf{f}'_i$ with $\Delta \mathbf{d}_i = \mathbf{d}-\mathbf{d}_i$ and input each concatenated feature into a small network to yield a blending weight~$w^c_i$.
The final color for this 5D location is blended via a soft-argmax operator~$\mathbf{c} =\sum_{i=1}^N [C_i\exp(w^c_i) / \sum_{j=1}^N \exp(w^c_j)]$. An alternative to predicting blending weights would be to directly regress $\mathbf{c}$. However, we found that direct regression of $\mathbf{c}$ led to decreased performance.

\subsection{Rendering and training}
The approach introduced in the previous section allows us to compute the color and density value at continuous 5D location. To render the color of a ray $\mathbf{r}$ passing through the scene, we first query the colors and densities of $M$ samples on the ray, and then accumulate colors along the ray modulated by densities:
\begin{align}
    \tilde{C}(\mathbf{r}) = \sum_{k=1}^{M} T_k (1 - \exp(-\sigma_k )) \mathbf{c}_k\,, \\
    \text{where} \quad T_k = \exp \bigg(-\sum_{j=1}^{k-1} \sigma_j \bigg).
\end{align}
Here samples from $1$ to $M$ are sorted to have ascending depth values. $\mathbf{c}_k$ and $\sigma_k$ denote the color and density of the $k$-th sample on the ray, respectively.

\medskip
\noindent \textbf{Hierarchical volume rendering.}
One benefit of having continuous RGB-$\sigma$ predictions is that it allows more adaptive and efficient sampling in space.
Following NeRF~\cite{nerf}, we perform hierarchical volume sampling and simultaneously optimize two networks, a coarse IBRNet and a fine IBRNet, with identical network architecture. At the coarse scale, we sample a set of $M_c$ locations at equidistant disparity (inverse depth) which results in equal intervals between adjacent point projections in pixel space. Given the prediction of the coarse network, we then conduct a more informed sampling of points along each ray, where samples are more likely to be located at relevant regions for rendering. We sample an additional $M_f$ locations and use all $M_c+M_f$ locations to render the fine results as in NeRF~\cite{nerf}. In our network, we set both $M_c$ and $M_f$ to $64$. 

\medskip
\noindent \textbf{Training objective.}
We render the color of each ray using both the coarse and fine set of samples, and minimize the mean squared error between the rendered colors and ground-truth pixel colors for training:
\begin{equation}
\mathcal{L}=\sum_{\mathbf{r}\in R}\left[\norm{\Tilde{C}_c(\mathbf{r}) - C(\mathbf{r})}_2^2 + \norm{\Tilde{C}_f(\mathbf{r}) - C(\mathbf{r})}_2^2 \right]
\label{eq:objective}
\end{equation}
where $R$ is the set of rays in each training batch.
This allows us to train the feature extraction network as well as the coarse and fine IBRNet simultaneously.

\medskip 
\noindent \textbf{Fine-tuning.}
Trained on large amounts of data, our method is able to generalize well to novel scenes. Our method can also be fine-tuned per scene using the same objective at training time~(Eq.~\ref{eq:objective}) but only on a specific scene, thereby improving synthesis performance on that scene. 

\subsection{Implementation details}
\noindent \textbf{Source and target view sampling.} Given multiple views of a scene, we construct a training pair of source and target view by first randomly selecting a target view, and then sampling $N$ nearby views as source views. To select source views, we first identify a pool of $n \cdot N$
nearby views~($n$ is sampled uniformly at random from $[1, 3]$), and then randomly sample $N$ views from the pool. This sampling strategy simulates various view densities during training and therefore helps the network generalize across view densities. During training, we sample $N$ uniformly at random from $[8, 12]$. 

\newcommand{\tablespace}{\,\,\,\,}
\newcommand{\halftablespace}{\,}
\setlength{\tabcolsep}{4pt}
\begin{table*}[t]
\centering
\begin{tabular}{l|c|ccc|ccc|ccc}
\toprule
& & \multicolumn{3}{c|}{Diffuse Synthetic $360^\circ$~\cite{sitzmann2019deepvoxels}} & \multicolumn{3}{c|}{Realistic Synthetic $360^\circ$~\cite{nerf}} & \multicolumn{3}{c}{Real Forward-Facing~\cite{mildenhall19}} \\
Method & Settings & PSNR$\uparrow$ & SSIM$\uparrow$ & LPIPS$\downarrow$ & PSNR$\uparrow$ & SSIM$\uparrow$ & LPIPS$\downarrow$ & PSNR$\uparrow$ & SSIM$\uparrow$ & LPIPS$\downarrow$ \\
\midrule
LLFF~\cite{mildenhall19}& \multirow{2}*{\makecell{No per-scene\\optimization}} & $34.38$ & $0.985$ & $0.048$ & $24.88$ & $0.911$ & $0.114$ & $24.13$ & $0.798$ & $0.212$\\
Ours & & $\mathbf{37.17}$ & $\mathbf{0.990}$ & $\mathbf{0.017}$ & $\mathbf{25.49}$ & $\mathbf{0.916}$ & $\mathbf{0.100}$ & $\mathbf{25.13}$ & $\mathbf{0.817}$ & $\mathbf{0.205}$\\
\midrule
SRN~\cite{srn} & \multirow{2}*{\makecell{\\ Per-scene \\ optimization}} & $33.20$ & $0.963$ & $0.073$ & $22.26$ & $0.846$ & $0.170$ & $22.84$ & $0.668$ & $0.378$ \\
NV~\cite{neuralvolumes} & & $29.62$ & $0.929$ & $0.099$ & $26.05$ & $0.893$ & $0.160$ & - & - & - \\

NeRF~\cite{nerf} & & $40.15$ & $0.991$ & $0.023$ & $\mathbf{31.01}$ & $\mathbf{0.947}$ & $0.081$ & $26.50$ & $0.811$ & $0.250$\\

Ours$_{\text{ft}}$ & & $\mathbf{42.93}$ & $\mathbf{0.997}$ & $\mathbf{0.009}$ & $28.14$ & $0.942$ & $\mathbf{0.072}$ & $\mathbf{26.73}$ & $\mathbf{0.851}$ & $\mathbf{0.175}$\\
\bottomrule
\end{tabular}
\vspace{-2mm}
\caption{\textbf{Quantitative comparison on datasets of synthetic and real images.} Our evaluation metrics are PSNR/SSIM (higher is better) and LPIPS~\cite{lpips} (lower is better). 
Both Ours and LLFF~\cite{mildenhall19} are trained on large amounts of training data and then evaluated on all test scenes without any per-scene tuning. Ours consistently outperforms LLFF~\cite{mildenhall19} on all datasets. We also compare our method with neural rendering methods~(SRN~\cite{srn}, NV~\cite{neuralvolumes}, and NeRF~\cite{nerf}) that train a separate network for each scene. To compete fairly with these methods, we also fine-tune our pretrained model on each scene~(Ours$_{\text{ft}}$). After fine-tuning, Ours$_{\text{ft}}$ is competitive with the state-of-the-art method NeRF~\cite{nerf}.
}%

\label{table:results}  
\end{table*}
\setlength{\tabcolsep}{1.4pt}

\medskip
\noindent \textbf{Network details.}
We implement the image feature extraction network using a U-Net like architecture with  ResNet34~\cite{he2016deep} truncated after \texttt{layer3} as the encoder, and two additional up-sampling layers with convolutions and skip-connections as the decoder. 
We decode two sets of feature maps in the final decoding layer, to be used as input to the coarse and fine IBRNet respectively. Both coarse and fine feature maps have $d=32$ dimensions, and are $1/4\times$ the original image size. For IBRNet, 
the dimension of the density feature $\mathbf{f}_\sigma$ is $16$ and we use $4$ heads for the self-attention module in ray transformer. Please refer to the supplementary material for more network details.

\medskip
\noindent\textbf{Training details.}
We train both the feature extraction network and the IBRNet end-to-end on datasets of multi-view posed images using Adam \cite{kingma2014adam}. The base learning rates for feature extraction network and IBRNet are $10^{-3}$ and $5 \times 10^{-4}$, respectively, which decay exponentially along with the optimization. During fine-tuning, we optimize both our 2D feature extractor and IBRNet itself using smaller base learning rates~($5 \times 10^{-4}$ and $2 \times 10^{-4}$).  
For pretraining, we train on eight V100 GPUs with a batch size of 3,200 to 9,600 rays depending on image resolution and the number of source views, which takes about a day to finish.

%% file: 04-experiment.tex
\section{Experiments}

\input{realresults.tex}

We evaluate our method in two ways: a) directly evaluating our pretrained model on all test scenes without any fine-tuning~(Ours). Note that we train only \emph{one} model and evaluate it on all test scenes. And b) fine-tuning our pretrained model on each test scene before evaluation~(Ours$_{\text{ft}}$), similar to NeRF.

\subsection{Experimental Settings}

\noindent\textbf{Training datasets.}
Our training data consists of both synthetic data and real data. For synthetic data, we generate object-centric renderings of the 1,023 models in Google Scanned Objects~\cite{google_scanned_objects}.
For real data, we use RealEstate10K~\cite{zhou2018stereo}, the Spaces dataset~\cite{flynn2019deepview}, and 102 real scenes from handheld cellphone captures~(35 from LLFF~\cite{mildenhall19} and 67 from ourselves). RealEstate10K is a large video dataset of indoor scenes with camera poses.
The Spaces dataset contains 100 scenes captured by a 16-camera rig.
Each of the cellphone captured scenes has 20-60 images captured by forward-facing cameras distributed roughly on 2D grids~\cite{mildenhall19}.  We use COLMAP~\cite{schonberger2016structure} to estimate camera parameters and scene bounds for our captures.
Our training data includes various camera setups and scene types, which allows our method to generalize well to unseen scenarios. 

\medskip
\noindent \textbf{Evaluation datasets}.
We use both synthetic rendering of objects and real images of complex scenes for evaluation, as in NeRF~\cite{nerf}. 
The synthetic data consists of four Lambertian objects with simple geometry from the DeepVoxels~\cite{sitzmann2019deepvoxels} dataset, and eight objects with complicated geometry and realistic materials from the NeRF synthetic dataset. 
Each object in DeepVoxels has 479 training views and 1,000 test views at $512\times512$ resolution sampled on the upper hemisphere. Objects in the NeRF synthetic dataset have 100 training views and 200 test views at $800\times800$ resolution sampled either on the upper hemisphere or the full sphere. The real dataset contains 8 complex real-world
scenes captured with roughly forward-facing images from NeRF. Each scene is captured with 20 to 62
images at $1008\times756$ resolution. $1/8$th of the images are held out for testing.

\medskip
\noindent \textbf{Baselines}. We compare our method against the current best view synthesis methods from two classes: \emph{MPI-based methods} and \emph{neural rendering methods}.  For MPI-based methods, we compare Ours with a state-of-the-art method LLFF~\cite{mildenhall19} which uses a 3D convolutional neural network to predict a discretized grid for each input image and blends nearby MPIs into the novel viewpoint. 
Both Ours and LLFF are designed to generalize to novel scenes. We compare Ours$_{\text{ft}}$ with neural rendering methods: Neural Volumes (NV)~\cite{neuralvolumes}, Scene Representation Networks (SRN)~\cite{srn}, and NeRF~\cite{nerf}. These methods train separate networks for each scene or each class of scenes, and either do not generalize at all, or only generalize within object categories.
When evaluating against these neural scene representation methods we fine-tune our model per-scene, as do the baselines.

\subsection{Results}
To render each test view we sample 10 source views from the training set for all evaluation datasets. 
We evaluate our method and our baselines using PSNR, SSIM, and LPIPS~\cite{lpips}. Results can be seen in Tab.~\ref{table:results} and in Fig.~\ref{fig:realresults}.

Tab.~\ref{table:results} shows that our pretrained model generalizes well to novel scenes and consistently outperforms LLFF~\cite{mildenhall19} on all test scenes. After fine-tuning, our model's performance is competitive with state-of-the-art neural rendering methods like NeRF~\cite{nerf}: We outperform NeRF on both \emph{Diffuse Synthetic $360^\circ$} and \emph{Real Forward-Facing}~\cite{mildenhall19} and only have lower PSNR and SSIM on \emph{Realistic Synthetic $360^\circ$}. 
One reason that we underperform NeRF on \emph{Realistic Synthetic $360^\circ$} is that this data has very sparse training views for scenes with very complex geometry. Therefore, the information contained in the limited set of local source views may be insufficient for synthesizing a novel view, in which case methods that optimize a global radiance field using all views like NeRF may be better suited.

On the \emph{Real Forward-Facing}~\cite{mildenhall19} data Ours$_\text{ft}$ achieves substantially better SSIM and LPIPS than NeRF~\cite{nerf}, indicating that our synthesized images look more photo-realistic. A key component in NeRF is the use of a positional encoding~\cite{tancik2020fourier} which helps generate high-frequency details. But positional encoding may also cause unwanted high-frequency artifacts in images~(see Fig.~\ref{fig:realresults} \textit{Orchid}), reducing perceptual quality. In contrast, our model does not use positional encoding to regress colors but instead blends colors from source views, biasing our synthesized images to look more like natural images. Our method produces reasonable proxy geometry and can be used to produce temporarily consistent videos. Please see the supplemental material for more detail.

\subsection{Ablations and analysis}
\noindent\textbf{Ablation studies}.
We conduct ablation studies~(Tab.~\ref{tab:ablation}) to investigate the individual contribution of key aspects of our method, using our pretrained models on \emph{Real Forward-Facing~\cite{mildenhall19}} data.
For ``No ray transformer'', we remove the ray transformer  so that the density prediction for samples on each ray is independent and local. Removing this module causes the network to fail to predict accurate densities, leading to ``black hole'' artifacts and blurriness~(see supplementary material for qualitative comparison). 
For ``No view directions'', we remove the view direction input in our network. This includes removing the weighted pooling in Eq.~\ref{eq:pooling_weight} and removing the viewing direction as a network input for color prediction. This reduces our model's ability to reproduce view-dependent effects such as specularities, but does not reduce performance as significantly as seen in NeRF because our model blends nearby input images (which themselves contain view-dependent effects). For ``Direct color regression'', we directly predict the RGB values at each 5D location instead of the blending weights. This is modestly worse than blending image colors from source views. 
 
\begin{table}[t]
    \centering
    \setlength{\tabcolsep}{6pt}
    \begin{tabular}{lccc}
    \toprule
            & PSNR$\uparrow$ & SSIM$\uparrow$ & LPIPS$\downarrow$ \\
            \midrule
        No ray transformer & $21.31$ & $0.675$ & $0.355$ \\
        No view directions & $24.20$ & $0.796$ & $0.243$\\
        Direct color regression & $24.73$ & $0.810$ & $0.220$ \\
        Full model Ours &  $\mathbf{25.13}$ & $\mathbf{0.817}$ & $\mathbf{0.205}$ \\ 
    \bottomrule
    \end{tabular}
    \vspace{-2mm}
    \caption{\textbf{Ablation study on \emph{Real Forward-Facing}~\cite{mildenhall19} data.} We report the metrics for the pretrained model of each ablation without per-scene fine-tuning.}
    \label{tab:ablation}
\end{table}

\begin{figure}[b]
    \centering
    \includegraphics[width=0.8\linewidth]{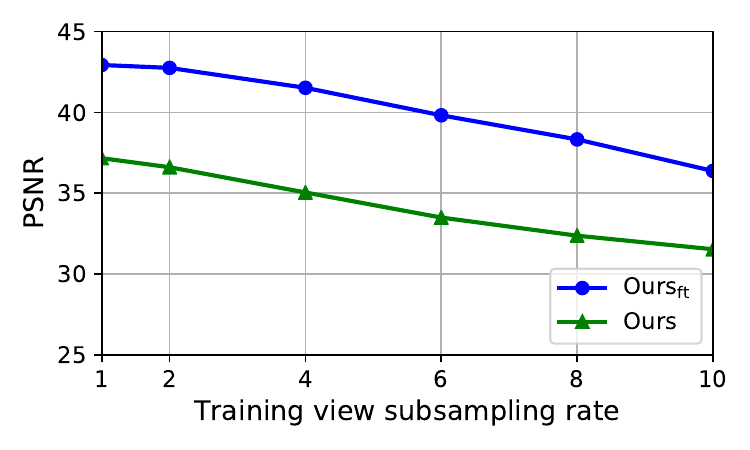}
    \vspace{-2mm}
    \caption{\textbf{Sensitivity to source view density}. We subsample the views provided in \emph{Diffuse Synthetic $360^\circ$}~\cite{sitzmann2019deepvoxels} by $\{2, 4, 6, 8, 10\}$ to create varying source view densities on the upper hemisphere. }
    \label{fig:view density}
\end{figure}

\medskip
\noindent \textbf{Sensitivity to source view density.}
We investigate how our method degrades as the source views become sparser on \emph{Diffuse Synthetic $360^\circ$}~\cite{sitzmann2019deepvoxels}. Each scene in the original data provides 479 training views randomly distributed on the upper hemisphere. We uniformly sub-sample the training views by factors of $2, 4, 6, 8, 10$ to create varying view densities. We report the PSNR for both the pretrained model Ours and the fine-tuned model Ours$_{\text{ft}}$ in Fig.~\ref{fig:view density}. Our method degrades reasonably as the input views become sparser.

\medskip
\noindent \textbf{Computational overhead at inference.}  Here we investigate the computation required by IBRNet in comparison to other methods.
Our inference pipeline for rendering a target view can be divided into two stages: We first extract image features from all source views of interest, which must only be done once as features can then be used repeatedly for rendering. Given these pre-computed features we use our proposed IBRNet to produce the colors and densities for samples along each camera ray. Because image features can be precomputed, the major computational cost at inference is incurred by IBRNet. 
Since our method interpolates novel views from a set of source views, the \#FLOPs of IBRNet depends on the size of the local working set, i.e., the number of source views~(\#Src.Views). We report the size of IBRNet and \#FLOPs required to render a single pixel under varying numbers of source views in Tab.~\ref{tab:inference_performance}. 

We see that IBRNet requires less than $20\%$ of the \#FLOPs than NeRF requires while achieving comparable performance. We consider two explanations for IBRNet's increased efficiency. First, NeRF uses a single MLP to encode the whole scene. To query any point in the scene, one must touch every parameter of the MLP. As the scene becomes more complex, NeRF needs more network capacity, which leads to larger \#FLOPs for each query. In contrast, our method interpolates nearby views locally, and therefore the \#FLOPs does not grow with the scale of the scene but only with the size of the local working set.
This property also makes our approach more dynamic, enabling applications like streaming large scenes and adjusting computational costs on-the-fly to suit a device.
Second, we encode the scene as posed images and features, a more structured and interpretable representation than the MLP used in NeRF. A simple projection operation allows us to fetch the most relevant information from source views for estimating radiance at a point in space. IBRNet must only learn how to combine that information, not how to synthesize it from scratch.

\begin{table}[t]
    \centering
    \setlength{\tabcolsep}{3pt}
    \resizebox{\linewidth}{!}{%
    \begin{tabular}{lcccccc}
    \toprule
     Method &  \#Params & \#Src.Views & \#FLOPs & PSNR$\uparrow$ & SSIM$\uparrow$ & LPIPS$\downarrow$\\
    \midrule
      SRN & 0.55M  & -  & 5M & $22.84$ & $0.668$ & $0.378$ \\
      NeRF & 1.19M  & -  & 304M & $26.50$ & $0.811$ & $0.250$ \\
    \midrule
    \multirow{3}*{
    Ours$_{\text{ft}}$
    } & \multirow{3}*{
    0.04M
    }
     & 5  & 29M  & $25.80$ & $0.828$  & $0.190$ \\
    & & 8  & 45M  & $26.56$ & $0.847$ & $0.176$ \\
    &  & 10  & 55M  & $\mathbf{26.73}$ & $\mathbf{0.851}$ & $\mathbf{0.175}$ \\
    \bottomrule
    \end{tabular}}
    \caption{
    \small
    \textbf{Network size and computational cost at inference.} The network size of our MLP is much smaller than SRN~\cite{srn} and NeRF~\cite{nerf}. All \#FLOPs reported are for rendering a single pixel. Both NeRF~\cite{nerf} and Ours$_{\text{ft}}$ use hierarchical volume sampling. $M_c$, $M_f$ are set to $64, 128$ respectively for NeRF, and $64, 64$ for Ours$_{\text{ft}}$. SRN~\cite{srn} uses ray marching with only 10 steps thus having much fewer \#FLOPs.
    For our method, the number of \#FLOPs scales roughly linearly with the number of source views~(\#Src.Views). 
    }
    \label{tab:inference_performance}
\end{table}

%% file: realresults.tex
\newcommand{\resultsfigheight}{0.9in}
\newcommand{\resultscropheight}{1.0in}

\newcommand{\croporchid}[1]{
  \makecell{
  \includegraphics[height=\resultscropheight]{#1} \\
  }
}

\newcommand{\croptrex}[1]{
  \makecell{
  \includegraphics[height=\resultscropheight]{#1}   \\
  }
}

\newcommand{\cropfern}[1]{
  \makecell{
  \includegraphics[height=\resultscropheight]{#1} \\
  }
}

\newcommand{\crophorns}[1]{
  \makecell{
  \includegraphics[height=\resultscropheight]{#1} \\
  }
}

\begin{figure*}[t]
\centering
\scriptsize
\begin{tabular}{@{}c@{\,\,}c@{}c@{}c@{}c@{}c@{}}
\makecell[c]{
\includegraphics[trim={0px 0px 0px 0px}, clip, height=\resultsfigheight]{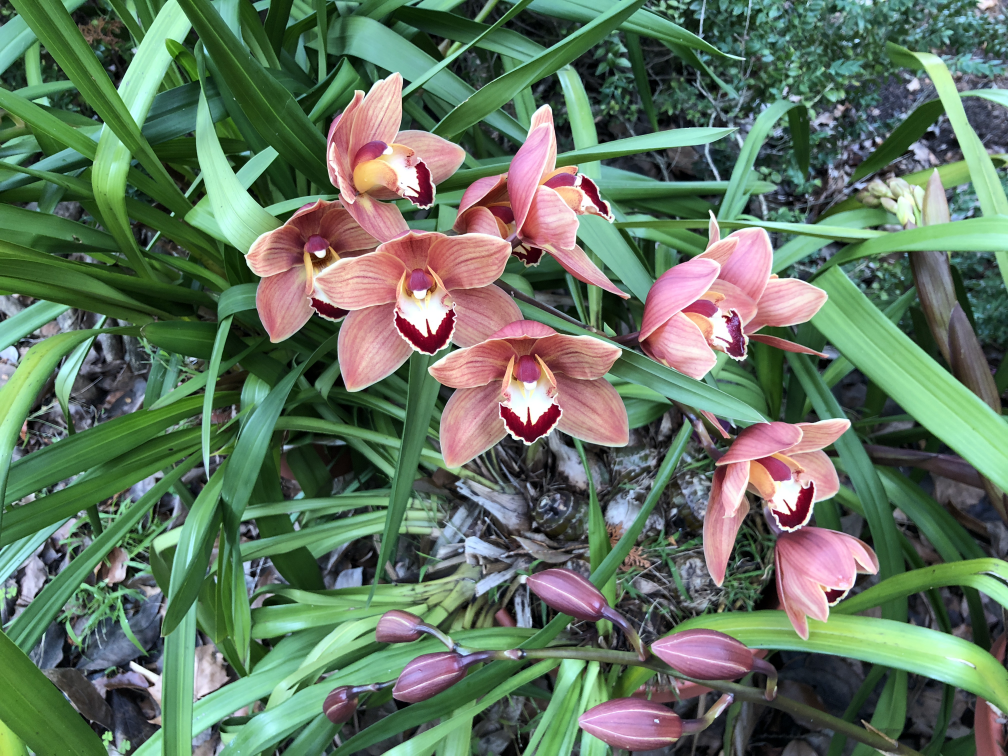}
\\
\textit{Orchid}
}
& 
\croporchid{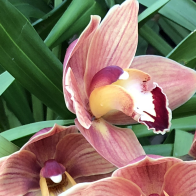} &
\croporchid{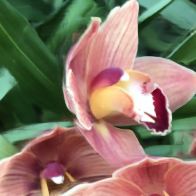} &
\croporchid{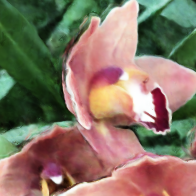} &
\croporchid{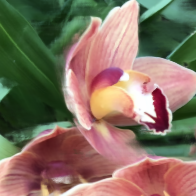} &
\croporchid{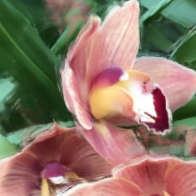} \\
\makecell[c]{
\includegraphics[trim={0px 0px 0px 0px}, clip, height=\resultsfigheight]{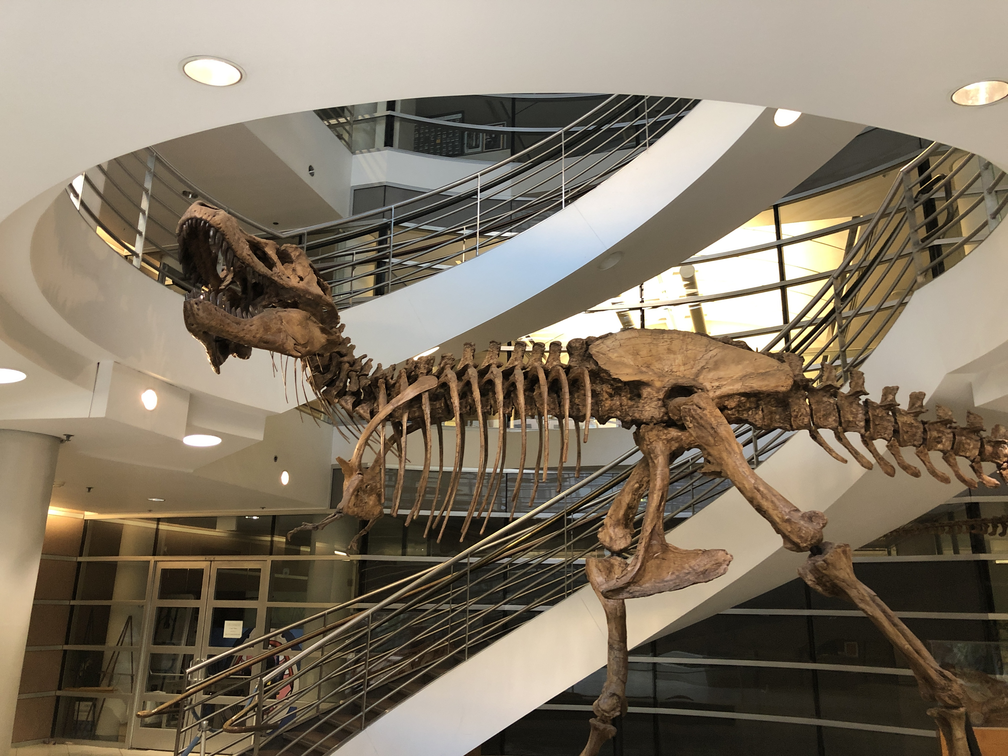}
\\
\textit{T-Rex}
}
& 
\croptrex{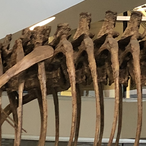} &
\croptrex{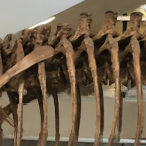} &
\croptrex{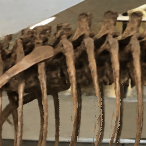} &
\croptrex{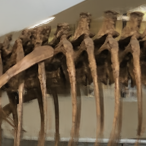} &
\croptrex{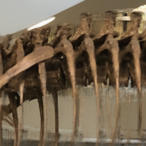} \\
\makecell[c]{
\includegraphics[trim={0px 0px 0px 0px}, clip, height=\resultsfigheight]{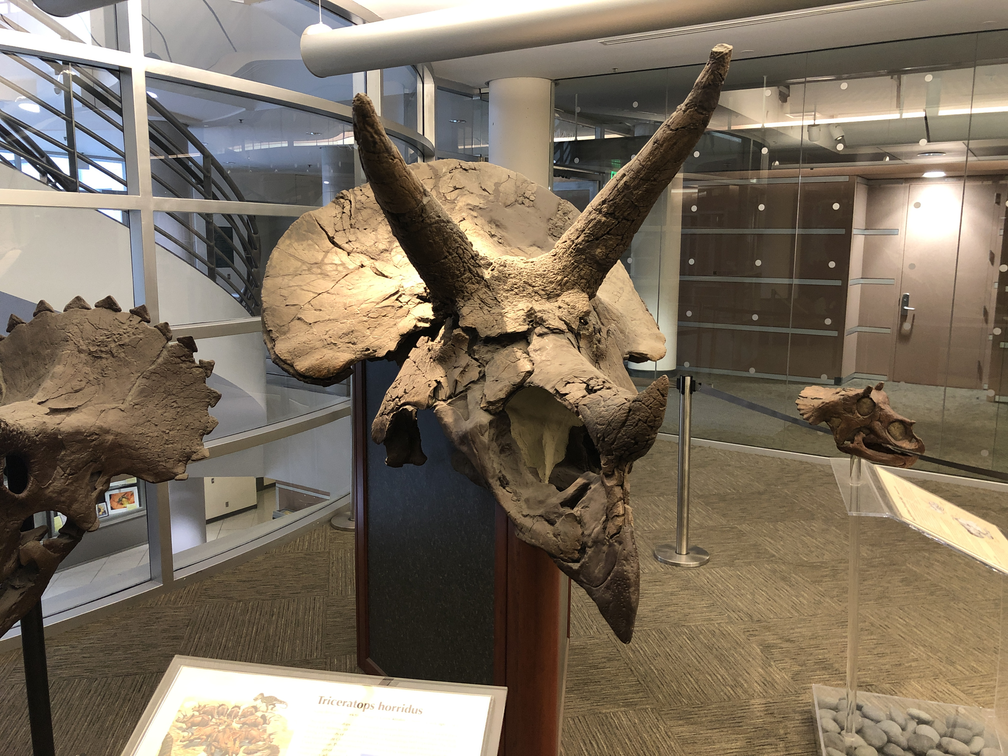}
\\
\textit{Horns}
}
& 
\crophorns{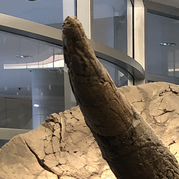} &
\crophorns{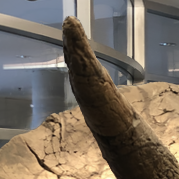} &
\crophorns{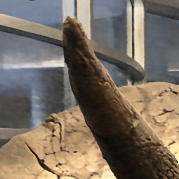} &
\crophorns{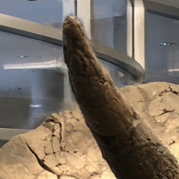} &
\crophorns{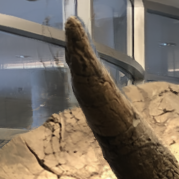} \\
\makecell[c]{
\includegraphics[trim={0px 0px 0px 0px}, clip, height=\resultsfigheight]{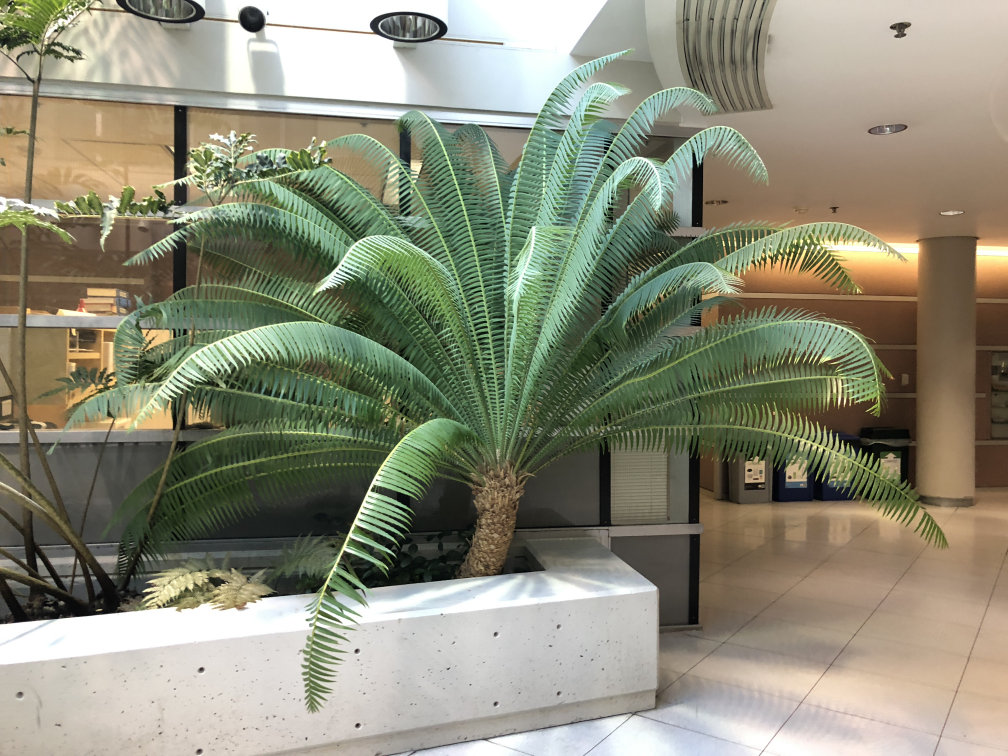}
\\
\textit{Fern}
}
& 
\cropfern{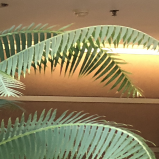} &
\cropfern{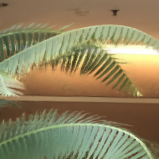} &
\cropfern{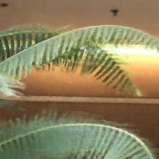} &
\cropfern{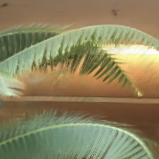} &
\cropfern{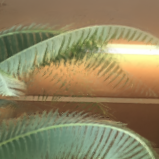} \\
& Ground Truth  &  Ours$_{\text{ft}}$ & NeRF~\cite{nerf} & Ours & LLFF~\cite{mildenhall19}
\end{tabular}
\caption{\textbf{Qualitative comparison on \emph{Real-Forward-Facing} data.} Our method can more accurately recover fine details in both geometry and appearance, and produce images that are perceptually more similar to ground-truth than other methods. LLFF~\cite{mildenhall19} has difficulty recovering clean and accurate boundary~(ghosting artifacts in \textit{Orchid} and repeated edges in \textit{Horns}), and fails to capture thin structures~(ribs in \textit{T-Rex}) or partially occluded origins~(leaves in \textit{Fern}). NeRF~\cite{nerf}-rendered images exhibits unrealistic noise in \textit{Orchid}. Meanwhile, the texture on the petals is missing and the region near the boundary of the petals are not well recovered. Our method is also slightly better than NeRF at fine structures in \textit{T-Rex} and \textit{Fern}, and can reconstruct more details of the reflection on the glass in \textit{Horns}.
}
\vspace{-0.5em}
\label{fig:realresults}
\end{figure*}

%% file: 05-conclusion.tex
\section{Conclusion}
We proposed a learning-based multi-view image-based rendering framework that synthesizes novel views of a scene by blending pixels from nearby images with weights and volume densities inferred by a network comprising an MLP and ray transformer.  This approach combines the advantages of IBR and NeRF to produce state-of-the-art rendering quality on complex scenes without requiring precomputed geometry~(unlike many IBR methods), storing expensive discretized volumes~(unlike neural voxel representations), or expensive training for each new scene (unlike NeRF).

\medskip
\noindent \textbf{Acknowledgements}
We thank Jianing Wei, Liangkai Zhang, Adel Ahmadyan, and Bhav Ashok for helpful discussions.

%% file: supp.tex
\clearpage
\section{Appendix}
\appendix

\section{Additional implementation details}
\noindent \textbf{Feature extraction network architecture.}
To render a target view, our system takes a set of source views and extracts their features using a network with shared weights. We implement the feature extraction network using a U-Net-like architecture, where the encoder is adapted from ResNet34~\cite{he2016deep} as implemented in PyTorch~\cite{paszke2017automatic}. We replace all Batch Normalization~\cite{ioffe2015batch} with Instance Normalization~\cite{ulyanov2016instance} as in~\cite{luo2020attention}, and remove max-pooling and use instead strided convolutions. Our network is fully convolutional and accepts input images of variable size. We take a single image of size $640 \times 480 \times 3$ as an example input and present a detailed network architecture in Tab.~\ref{tab:feature_net}. Our code and model will be made available.

\begin{table}[h]
    \centering
    \setlength{\tabcolsep}{5pt}
    \resizebox{\linewidth}{!}{%
    \begin{tabular}{lll}
        \toprule
        Input~(\textbf{id}: dimension)& Layer & Output~(\textbf{id}: dimension)\\
        \midrule
        \textbf{0}: $640\times480\times3$ & $7\times7$ Conv, $64$, stride $2$ & \textbf{1}: $320\times240\times64$\\
        \textbf{1}: $320\times240\times64$ & Residual Block 1 & \textbf{2}: $160\times120\times64$ \\
        \textbf{2}: $160\times120\times64$ & Residual Block 2 & \textbf{3}: $80\times60\times128$ \\
        \textbf{3}: $80\times60\times128$ & Residual Block 3 & \textbf{4}: $40\times30\times256$ \\
        \textbf{5}: $40\times30\times256$ & $3\times3$ Upconv, $128$, factor $2$ & \textbf{6}: $80\times60\times128$\\
        \text{[\textbf{3}, \textbf{6}]}: $80\times60\times256$ & $3\times3$ Conv, $128$ & \textbf{7}: $80\times60\times128$\\
        \textbf{7}: $80\times60\times128$ & $3\times3$ Upconv, $64$, factor $2$ & \textbf{8}: $160\times120\times64$ \\
        \text{[\textbf{2}, \textbf{8}]}: $160\times120\times128$ & $3\times3$ Conv, $64$ & \textbf{9}: $160\times120\times64$ \\
        \textbf{9}: $160\times120\times64$ & $1\times1$ Conv, $64$ & \textbf{Out}: $160\times120\times64$ \\ 
        \bottomrule
    \end{tabular}}
    \caption{\textbf{Feature extraction network architecture}. `Conv'' stands for a sequence of operations: convolution, rectified linear units (ReLU) and Instance Normalization~\cite{ulyanov2016instance}. ``Upconv'' stands for a bilinear upsampling with certain factor, followed by a ``Conv'' operation with stride 1. ``$\text{[}\cdot, \cdot\text{]}$'' represents channel-wise concatenation of two feature maps.
    The residual blocks have similar structures to those in the original ResNet34~\cite{he2016deep} design, except that the first residual block has stride equal to $2$ and all Batch Normalization layers are replaced with Instance Normalization layers. The output $64$-dimensional feature map will be split into two feature maps of $32$ dimension, which are then used as inputs to the coarse and fine IBRNet, respectively.}
    \label{tab:feature_net}
\end{table}
\medskip

\begin{figure*}
    \centering
    \includegraphics[width=\linewidth]{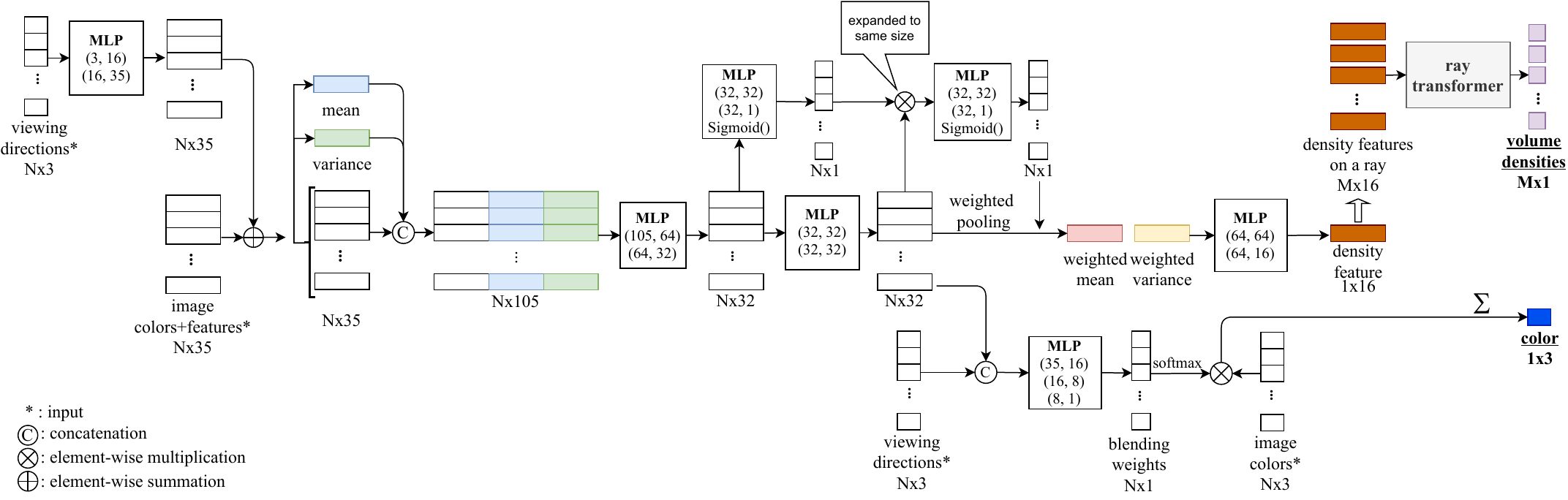}
    \caption{\textbf{IBRNet architecture}. $N$ denotes the number of source views and $M$ is the number of samples on a ray. 
    For each ``MLP'' box, $(i, j)$ represents a linear layer with input dimension $i$ and output dimension $j$. ELU is used between each two adjacent linear layers as the non-linear activation function. When ``MLP'' takes in a stack of feature vectors~(i.e., feature vectors from all source views), the MLP is applied to each feature vector with shared weights. ``weighted pooling'' computes the weighted mean and variance of the $N\times 32$ feature vector using the learned $N\times 1$ weight vector.
    The ray transformer module contains a single multi-head self-attention layer with the number of heads set to $4$. Viewing directions shown in the figures are relative viewing directions, i.e., the viewing direction of the query ray relative to the viewing directions from source views.  
    }
    \label{fig:ibr_net}
\end{figure*}

\noindent\textbf{IBRNet network architecture.}
Fig.~\ref{fig:ibr_net} shows the detailed network architecture of IBRNet and the process of predicting the volume density and color at a single 5D location. To obtain the density at each 5D location, we first aggregate the multi-view features drawn from source views to obtain a density feature that encodes the density information for this point. We then aggregate the density information of all samples on the ray to enable visibility reasoning for better density prediction. We implement this ray-wise operation using our proposed module called ``ray transformer''. 
Specifically, we first apply positional encoding~\cite{vaswani2017attention} to all density features on the ray, so that the network is aware of the spatial ordering for samples on the ray. We then use a single multi-head self-attention~\cite{vaswani2017attention} layer to incorporate long-range contextual information. Separately, to obtain the view-dependent color, we predict a set of blending weights to blend the image colors drawn from source views. IBRNet is invariant to permutations of source views, and supports a variable number of source views as well as samples on a ray.

\begin{figure*}[h]
    \centering
    \includegraphics[width=0.97\linewidth]{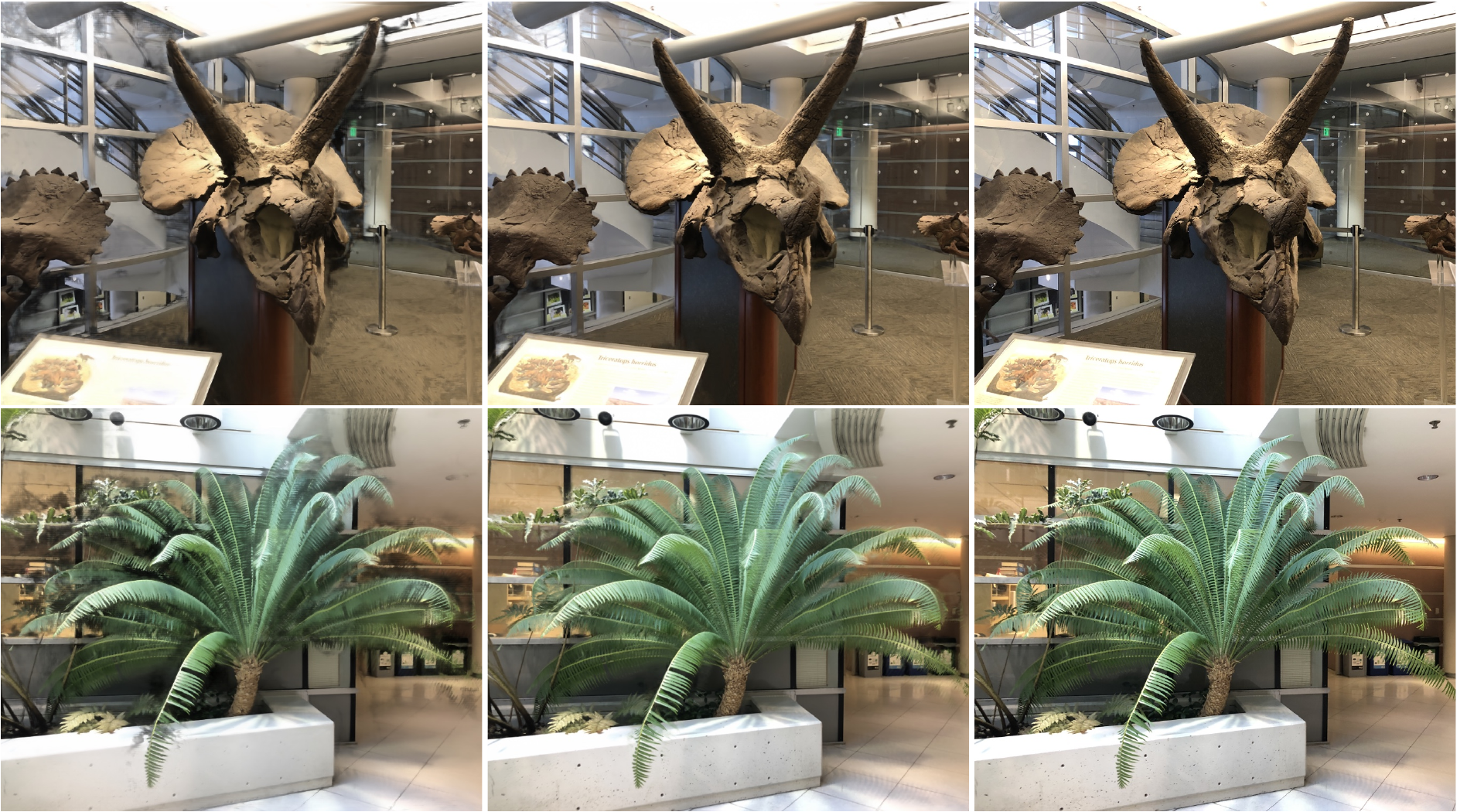}
    \caption{\textbf{Qualitative comparison of our model trained with and without ray transformer.} The first and second columns show the results of our pretrained model without and with the ray transformer module, respectively. The last column shows the ground truth images. Without ray transformer, the synthesized images exhibit severe ``black hole'' artifacts especially near occlusion boundaries where the network fails to infer surface locations correctly. Ray transformer eliminates such artifacts by enabling more informed density prediction.}
    \label{fig:ray_transformer_results}
\end{figure*}

\medskip
\noindent\textbf{Training details.}
At training time, we sample points in space and project them in the source views to fetch the corresponding colors and image features. However, the projected pixel for a sample may be located outside of the image plane. In this case, we discount this source view for this sample. If a point is not projected into the image plane of any source view, we set the volume density of this point to be zero. If there are less than 3 samples on a ray that have valid density values, we ignore this ray in the loss function during training. 

For pre-training, We train on eight V100 GPUs with a batch size of 3,200 to 9,600 rays depending on image resolution and the number of source views. Within each batch, we sample rays from eight different scenes. Within each scene, we sample rays randomly from a single image for training. Pre-training takes about a day to finish. For fine-tuning, the time needed is scene-dependent. It takes us about $6$ hours on a single V100 GPU to achieve the reported performance for each of the \emph{Real Forward-Facing}~\cite{mildenhall19} scenes. The fine-tuning stage may be accelerated with better hyperparameters for optimization.

\section{Additional qualitative results}
\noindent \textbf{Ray transformer}. We provide qualitative comparison of our model trained with and without ray transformer in Fig.~\ref{fig:ray_transformer_results}. Fig.~\ref{fig:ray_transformer_results} shows that our proposed ray transformer significantly improves the synthesis quality especially for challenging regions~(e.g., near occlusion boundaries).

\begin{figure*}[h]
    \centering
    \includegraphics[width=\linewidth]{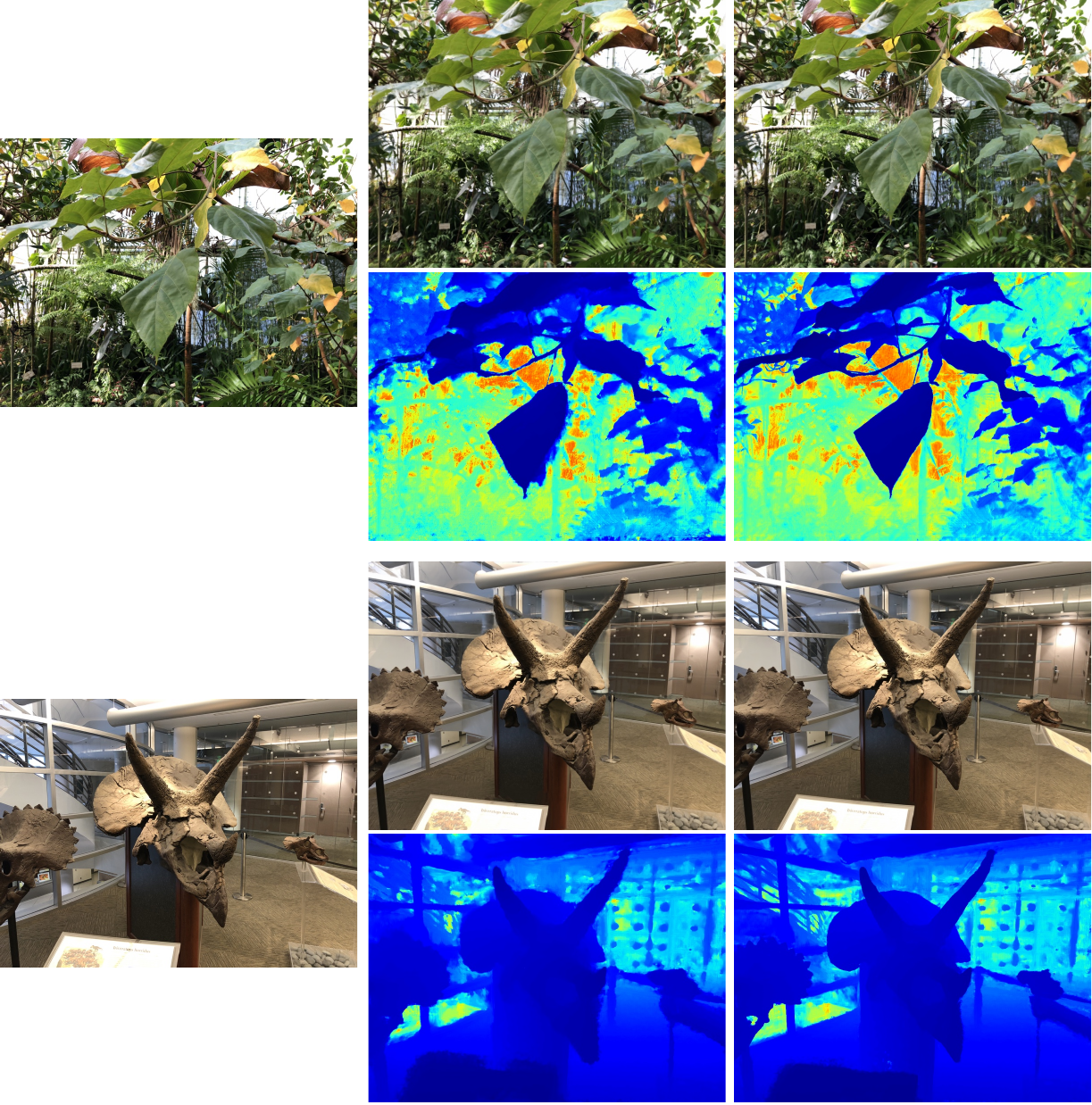}
    \caption{\textbf{Geometry Visualization}. We visualize the proxy geometry and synthesized images generated by our pretrained and fine-tuned models for two scenes \emph{leaves} and \emph{horns}. For each scene, the first column shows the ground truth image. The second column shows the results, i.e., synthesized image~(top) and depth map~(bottom), using our pretrained model. the last column shows the results of our model after finetuned on each scene.
    }
    \label{fig:depth_vis}
\end{figure*}

\medskip
\noindent\textbf{Geometry Visualization.}
We visualize the proxy geometry by accumulating the predicted density values on each ray. Fig.~\ref{fig:depth_vis} shows that our pretrained model produces reasonable proxy geometry, and the quality of the synthesized images and the underlying geometry improves when our model is fine-tuned.

\begin{figure*}
    \centering
    \includegraphics[width=0.97\linewidth]{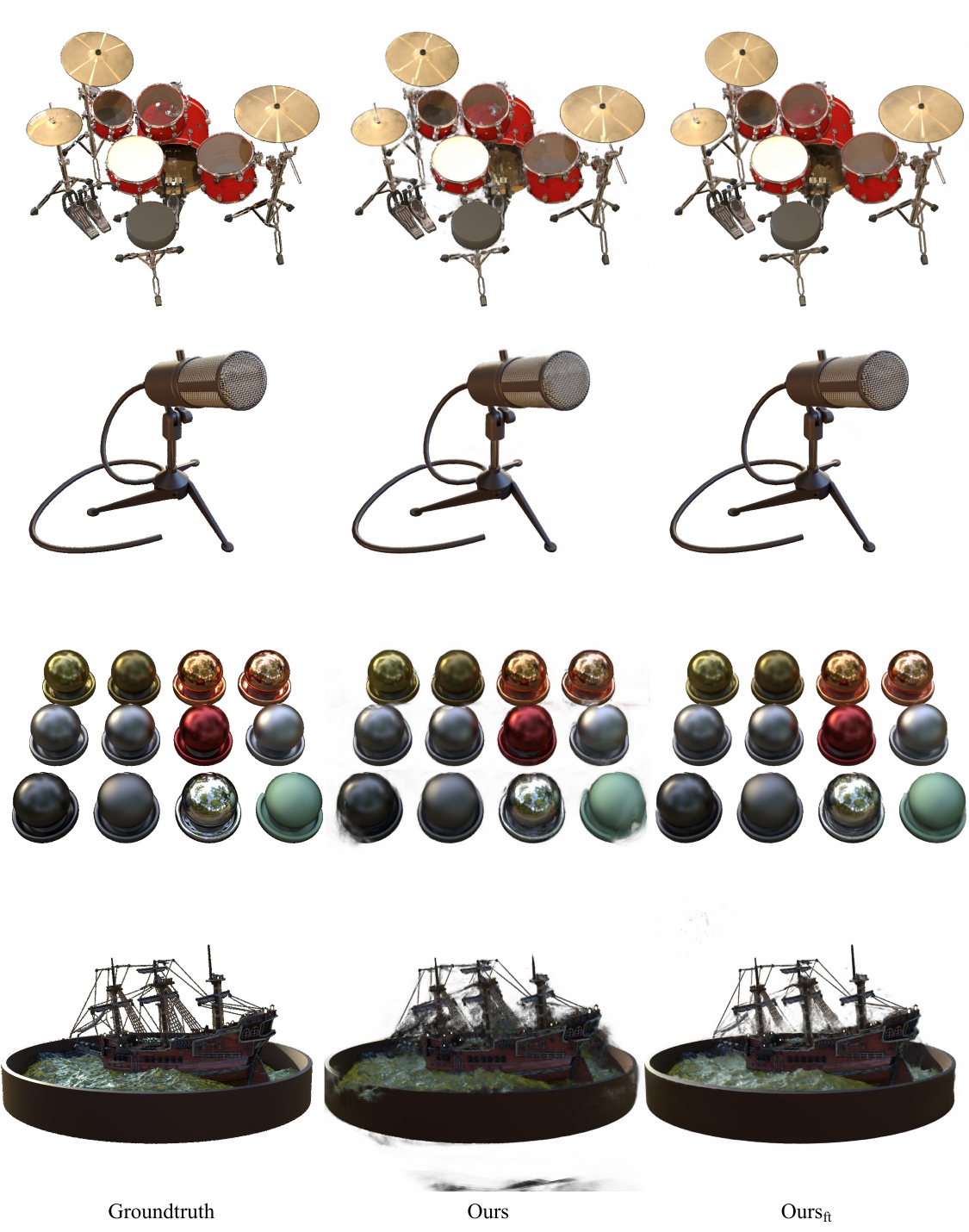}
    \caption{\textbf{Qualitative results on \emph{Realistic Synthetic $360^\circ$}~\cite{nerf}.} The first column shows the ground truth images. The second column shows the synthesized images without per-scene fine-tuning. The last column shows the synthesized images with per-scene fine-tuning. The last row is a failure case due to sparse source views and complex geometry.}
    \label{fig:qualitative_nerf_syn}
\end{figure*}

\medskip
\noindent \textbf{Qualitative results on \emph{Realistic Synthetic $360^\circ$~\cite{nerf}}.}
We provide qualitative results of our pretrained and finetuned model on \emph{Realistic Synthetic $360^\circ$~\cite{nerf}} in Fig.~\ref{fig:qualitative_nerf_syn}. The last column in Fig.~\ref{fig:qualitative_nerf_syn} shows a challenging scenario where our method does not work well.

